%% file: main.tex
\documentclass[sigconf]{acmart} 

\AtBeginDocument{%
  }

\copyrightyear{2025}
\acmYear{2025}
\setcopyright{acmlicensed}
\acmConference[MM '25] {Proceedings of the 33rd ACM International Conference on Multimedia}{October 27--31, 2025}{Dublin, Ireland.}
\acmBooktitle{Proceedings of the 33rd ACM International Conference on Multimedia (MM '25), October 27--31, 2025, Dublin, Ireland}
\acmISBN{979-8-4007-2035-2/2025/10}
\acmDOI{10.1145/3746027.3755845}

\usepackage{multirow}
\usepackage{colortbl}
\usepackage{hhline}
\usepackage{subcaption}

\definecolor{red}{RGB}{255,0,0}
\definecolor{green}{RGB}{0,255,0}
\definecolor{blue}{RGB}{0,0,255}
\definecolor{myblue}{RGB}{65, 105, 225}
\definecolor{orange}{RGB}{255,165,0}
\definecolor{title_gray}{gray}{.9}

\newcommand{\thickhline}{\noalign{\hrule height 1pt}}

\begin{document}

\title[RATopo: Improving Lane Topology Reasoning via Redundancy Assignment]{RATopo: Improving Lane Topology Reasoning via \\Redundancy Assignment}


\author{Han Li}
\email{lihan0620@buaa.edu.cn}
\orcid{0000-0001-9368-826X}
\affiliation{
  \department{School of Artificial Intelligence}
  \institution{Beihang University}
  \city{Beijing}
  \country{China}
}
\affiliation{
  \institution{Zhongguancun Academy}
  \city{Beijing}
  \country{China}
}

\author{Shaofei Huang}
\email{nowherespyfly@gmail.com}
\orcid{0000-0001-8996-9907}
\affiliation{
  \department{Faculty of Science and Technology}
  \institution{University of Macau}
  \city{Macau}
  \country{China}
}

\author{Longfei Xu}
\email{21371047@buaa.edu.cn}
\orcid{0009-0009-0750-3325}
\affiliation{
  \department{School of Computer Science and Engineering}
  \institution{Beihang University}
  \city{Beijing}
  \country{China}
}

\author{Yulu Gao}
\email{gyl97@buaa.edu.cn}
\orcid{0000-0002-3895-1288}
\affiliation{
  \department{School of Artificial Intelligence}
  \institution{Beihang University}
  \city{Beijing}
  \country{China}
}
\affiliation{
  \department{Hangzhou International Innovation Institute}
  \institution{Beihang University}
  \city{Hangzhou}
  \country{China}
}

\author{Beipeng Mu}
\email{mubeipeng@meituan.com}
\orcid{0009-0000-9240-7964}
\affiliation{
  \institution{Meituan}
  \city{Beijing}
  \country{China}
}

\author{Si Liu}
\authornote{Corresponding author.}
\email{liusi@buaa.edu.cn}
\orcid{0000-0002-9180-2935}
\affiliation{
  \department{School of Artificial Intelligence}
  \institution{Beihang University}
  \city{Beijing}
  \country{China}
}

\renewcommand{\shortauthors}{Han Li et al.}

\begin{abstract}

Lane topology reasoning plays a critical role in autonomous driving by modeling the connections among lanes and the topological relationships between lanes and traffic elements.
Most existing methods adopt a \textit{first-detect-then-reason} paradigm, where topological relationships are supervised based on the one-to-one assignment results obtained during the detection stage. This supervision strategy results in suboptimal topology reasoning performance due to the limited range of valid supervision.
In this paper, we propose \textbf{RATopo}, a \textbf{R}edundancy \textbf{A}ssignment strategy for lane \textbf{Topo}logy reasoning that enables quantity-rich and geometry-diverse topology supervision.
Specifically, we restructure the Transformer decoder by swapping the cross-attention and self-attention layers. This allows redundant lane predictions to be retained before suppression, enabling effective one-to-many assignment.
We also instantiate multiple parallel cross-attention blocks with independent parameters, which further enhances the diversity of detected lanes.
Extensive experiments on OpenLane-V2 demonstrate that our RATopo strategy is model-agnostic and can be seamlessly integrated into existing topology reasoning frameworks, consistently improving both lane-lane and lane-traffic topology performance 
(\textit{e.g.}, + 15.7\% and + 9.5\% on TopoLogic for $\text{TOP}_{ll}$ and $\text{TOP}_{lt}$ on OpenLane-V2 \textit{subset\_B}, respectively). The codes are released
at https://github.com/homothetic/RATopo.
\end{abstract}


\begin{teaserfigure}
  \includegraphics[width=\textwidth]{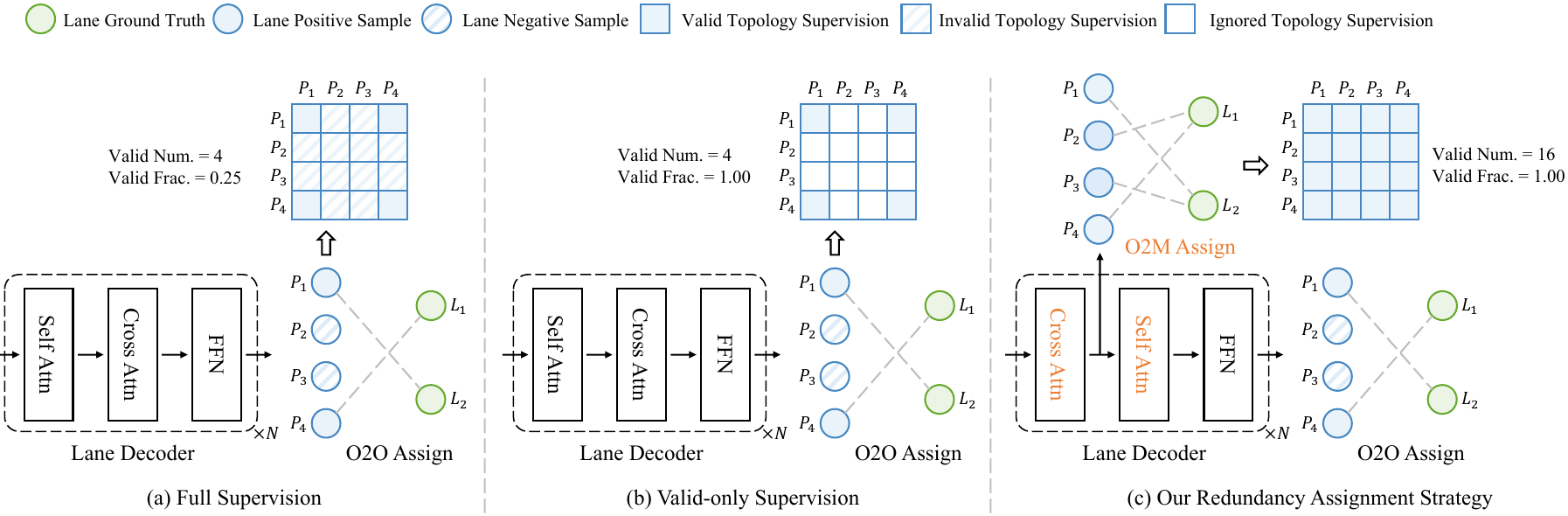}
  \caption{Comparison between different types of topology supervision. (a) Full supervision on both valid and invalid topologies, resulting in severe dominance of invalid cases. (b) Supervision on valid topologies only, which still suffers from sparse supervision of limited valid topologies resulting from one-to-one assignment. (c) Our redundancy assignment strategy increases valid topology supervision through a tailored one-to-many assignment scheme.}
  \Description{}
  \label{fig:teaser}
\end{teaserfigure}


\maketitle
   
\input{sec/1_introduction}
\input{sec/2_related_work}
\input{sec/3_method}
\input{sec/4_experiments}
\input{sec/5_conclusion}

\begin{acks}
This research is supported in part by National Key R\&D Program of China (2022ZD0115502), National Natural Science Foundation of China (NO.62461160308, U23B2010), "Pioneer" and "Leading Goose" R\&D Program of Zhejiang (No.2024C01161), Zhongguancun Academy Project (No.20240304), and Meituan.
\end{acks}

\bibliographystyle{ACM-Reference-Format}
\bibliography{main}


\end{document}

%% file: sec/1_introduction.tex
\section{Introduction} 
\label{sec:intro}
Lane topology reasoning has emerged as a critical component of autonomous driving perception to adapt to complex road structures and diverse driving conditions.
Building upon 3D lane detection~\cite{3dlanenet,genlanenet,Anchor3DLane}, it provides structured understanding of the driving scene to support advanced functions such as planning and control~\cite{UniAD, MP3, liang2020learning}.
Beyond perceiving individual lanes and traffic elements, lane topology reasoning further requires inferring their topological relationships. These include the connectivity between lanes (\textit{i.e.}, lane-lane topology) and the correspondence between lanes and traffic elements (\textit{i.e.}, lane-traffic topology), which together form a comprehensive scene-level topology graph.

Given the inherent dependency between detection and reasoning, most lane topology reasoning methods~\cite{TopoNet, TopoMLP, TopoLogic, Topo2D, LaneSegNet, SMERF} follow a \textit{first-detect-then-reason} paradigm.
Concretely, the detection stage employs DETR-style~\cite{DETR} decoders to obtain features and positions of detected instances, while the reasoning stage predicts their topological relationships according to their feature similarity and spatial distance.
Under such formulation, the training of the reasoning stage is tightly coupled with the label assignment results of the detection stage. 
Taking lane-lane topology reasoning as an example, one-to-one matching is typically performed between detected and ground-truth lanes. Topology relationships between ground-truth lanes (\textit{e.g.}, connected or unconnected) are then projected onto their corresponding positive lane predictions to obtain the supervision signals for lane-lane topology learning.

As shown in Figure~\ref{fig:teaser}(a), early works~\cite{TopoNet, TopoLogic} apply full supervision to all topology predictions, including both \textit{valid} ones composed of two positive lane predictions and \textit{invalid} ones involving at least one negative prediction.
However, this supervision strategy leads to a severe dominance of invalid topologies, with valid ones accounting for only a small fraction of all supervised ones (\textit{e.g.}, 4 out of 16 in the simplified example shown in Figure~\ref{fig:teaser}(a)).
This imbalance significantly hinders the model’s ability to learn meaningful topological relationships.
Later methods~\cite{TopoMLP, Topo2D} mitigate this issue by applying supervision to valid topologies only, as shown in Figure~\ref{fig:teaser}(b).
Although this improves the quality of supervision by eliminating ambiguous invalid cases, it does not increase the number of valid ones being supervised.
Given the inherent ambiguity of lane detection, where a single lane may correspond to multiple plausible geometric variants (\textit{e.g.}, unmarked intersections where multiple turning trajectories are equally plausible), many potential topological relationships involving these alternative predictions remain unexplored under the one-to-one assignment scheme.
We argue that the scarcity of valid topologies limits the model’s ability to capture the structural diversity required for both lane-lane and lane-traffic topology reasoning in complex driving scenes.

To overcome the limitations of prior topology supervision strategies relying on \textbf{one-to-one assignment}, we propose a simple yet effective \textbf{R}edundancy \textbf{A}ssignment strategy for lane \textbf{Topo}logy training~(\textbf{RATopo}) to increase valid topology supervision through a meticulous \textbf{one-to-many assignment} scheme.
As illustrated in Figure~\ref{fig:teaser}(c), simply increasing one positive prediction per lane raises the number of valid lane-lane topologies from 4 to 16.
However, due to the redundancy suppression mechanism in standard DETR-style~\cite{DETR} Transformer~\cite{transformer} decoders, \textit{naively applying one-to-many assignment proves ineffective}, as evidenced in Table~\ref{tab:one2many}. 
In particular, the combination of self-attention and one-to-one assignment implicitly suppresses redundant predictions, typically yielding only one high-quality prediction per ground-truth lane.
As a result, low-quality predictions are introduced, harming topology learning.
Therefore, to support effective one-to-many assignment, our RATopo strategy first restructures the Transformer~\cite{transformer} decoder by swapping the order of the cross-attention (CA) and self-attention (SA) layers (Figure~\ref{fig:teaser}(c)).
This reordering allows the CA layer to generate redundant lane predictions early, before they are suppressed by the SA layer, thereby preserving multiple high-quality variants per ground-truth lane for valid topology construction.
To further enhance the diversity of these lane predictions, we instantiate multiple parallel CA blocks with independent parameters, enabling the same set of lane queries to attend to diverse spatial and semantic patterns from the image features. 
Together, these two components allow RATopo to provide quantity-rich and geometry-diverse supervision signals for topology learning. 
Notably, our RATopo is model-agnostic and can be seamlessly integrated into various DETR-style topology frameworks, consistently yielding improvements for both lane-lane and lane-traffic topology prediction.

Compared with existing one-to-many assignment strategies in object detection, such as Group DETR~\cite{GroupDETR} and $\mathcal{H}$-DETR~\cite{HDETR}, our approach differs in two key aspects. 
First, existing strategies adopt one-to-many assignment by introducing additional query groups, which are primarily designed to enhance detection performance rather than facilitate topology reasoning. 
In contrast, our RATopo strategy is explicitly tailored for topology reasoning. 
It decouples the learning objectives of lane detection and topology reasoning by maintaining one-to-one assignment for detection to ensure sparse lane predictions, while leveraging multiple variants from the reordered CA layer to enrich valid topology supervision.
Second, although existing strategies can also be adapted to topology reasoning by adding additional multiple
groups of lane queries to obtain more valid
topologies, they merely increase the quantity of supervision without improving its density within the original query set.
Our RATopo strategy instead focuses on enhancing the diversity of the existing lane queries, enabling a larger proportion of lane predictions to be involved in topology loss computation.

Our main contributions are summarized as follows:
\begin{itemize}
    \item We identify the critical limitation of sparse valid topology supervision in existing lane topology reasoning methods, which stems from the inherent conflict between one-to-one label assignment in DETR-style~\cite{DETR} lane detectors and the geometric ambiguity of lane structure representations.
    \item We propose RATopo, a model-agnostic redundancy assignment strategy that breaks the one-to-one assignment bottleneck by restructuring the Transformer~\cite{transformer} decoder and introducing parallel cross-attention, enabling dense and geometrically diverse valid topology supervision.
    \item Extensive experiment results on OpenLane-V2~\cite{OpenLaneV2} show that our RATopo obtains consistent improvements on various topology reasoning methods (\textit{e.g.}, + 15.7\% and + 9.5\% on TopoLogic~\cite{TopoLogic} for lane-lane topology and lane-traffic topology on OpenLane-V2 \textit{subset\_B}, respectively), demonstrating its effectiveness and generality.
\end{itemize}

%% file: sec/2_related_work.tex
\section{Related Work}

\subsection{Lane Detection}
\noindent 
Currently, 2D lane detection methods can be roughly divided into four primary categories: anchor-based~\cite{laneatt, clrnet}, parameter-based~\cite{polylanenet, lstr}, 
segmentation-based~\cite{SCNN, UFSA},
and keypoint-based~\cite{ganet, fololane, RCLane}.  
Anchor-based methods, such as CLRNet~\cite{clrnet}, are currently mainstream and typically utilize equidistant 2D points as anchors, initially detecting lanes using high-level semantic features and subsequently refining them with low-level features.
3D lane detection extends 2D lane detection by estimating the 3D coordinates of lane.
Early methods~\cite{3dlanenet, 3dlanenet+, genlanenet, CLGo} simplify the projection process by assuming a flat ground plane, enabling the transformation of 2D image features into BEV features.
However, these methods suffer from significant errors in real-world driving scenarios, particularly on uphill or downhill roads.
To address this limitation, PersFormer~\cite{Persformer} adopts deformable attention~\cite{DeformableDETR} to generate robust BEV features. 
Recent advancements~\cite{curveformer,Anchor3DLane,Anchor3DLane++,Latr} have shifted the focus toward directly predicting 3D lane coordinates from front-view image features. 
For example, Anchor3DLane~\cite{Anchor3DLane, Anchor3DLane++} defines lane anchors in 3D space, projects them onto front-view image features to obtain more accurate anchor features for lane regression.

\subsection{Online HD Map Construction}
\noindent Constructing online high-definition (HD) maps from multi-view images involves the detection of static scene elements, such as pedestrian crossings, lane dividers, and road boundaries. 
Current methodologies can be categorized into two paradigms based on their output representations: rasterized HD map construction~\cite{vpn, gkt, cvt, BEVFormer, lss} and vectorized HD map construction~\cite{HDMapNet, VectorMapNet, MapTR, MapTRv2, MapQR, streammapnet, pivotnet}.
MapTR~\cite{MapTR} and MapTRv2~\cite{MapTRv2} represent map elements as point sets with equivalent permutations, incorporating a hierarchical query embedding scheme to efficiently encode structured map information.
MapQR~\cite{MapQR} introduces a scatter-and-gather query design, which scatters instance queries into hierarchical queries through position embedding, thereby reducing computation in the self-attention mechanism.
StreamMapNet~\cite{streammapnet} employs multi-point attention and temporal information to construct large-range and long-sequence local HD maps.

\subsection{Lane Topology Reasoning}
\noindent 
The introduction of the OpenLane-V2~\cite{OpenLaneV2} dataset has significantly advanced research on lane topology reasoning.
TopoNet~\cite{TopoNet} utilizes a scene graph neural network to effectively model the lane topology and enhance feature interaction within the lane detection network.
TopoMLP~\cite{TopoMLP} develops a simple MLP-based framework to interpret the lane topological relationships.
Topo2D~\cite{Topo2D} enhances 3D lane detection and topology reasoning using 2D lane priors. 
TopoLogic~\cite{TopoLogic} tackles the challenge of endpoint shifts in geometric space and introduces an interpretable method for lane topology reasoning based on lane geometric distance.
LaneSegNet~\cite{LaneSegNet} proposes lane segment perception as an innovative map learning framework, integrating a lane attention module to effectively capture long-range dependencies.

We propose a model-agnostic redundancy assignment strategy for lane topology reasoning, 
aiming to provide quantity-rich and geometry-diverse supervision signals for valid topology learning.
Our strategy can be seamlessly integrated into existing DETR-style~\cite{DETR} methods and results in significant performance improvements consistently.

\begin{figure*}[t]
  \centering
  \includegraphics[width=0.9\linewidth]{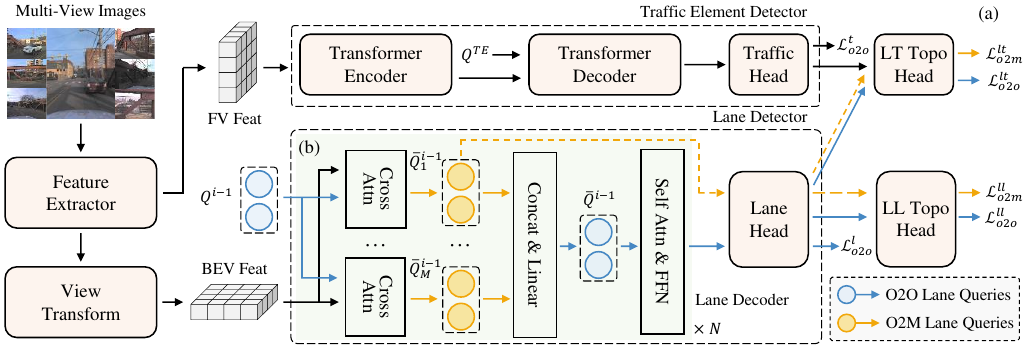}
  \caption{(a) A typical topology reasoning pipeline, with our RATopo strategy applied to each Transformer~\cite{transformer} layer of the lane decoder. (b) Lane decoder with redundancy assignment. \textit{The dashed arrows indicate data flow that exists only during training.}}
  \Description{}
  \label{fig:pipeline}
\end{figure*}

\subsection{One-to-many Assignment}
\noindent The one-to-many assignment strategy, which involves designating multiple predictions as positive samples for each ground truth, has proven effective in detection methods like Faster R-CNN~\cite{FasterRCNN} and FCOS~\cite{FCOS}.
For one-to-many assignment in DETR-style methods~\cite{DETR, DeformableDETR, ConditionalDETR, DABDETR, DINO}, Group DETR~\cite{GroupDETR} initializes multiple groups of object queries and performs self-attention and one-to-one assignment within each group, while $\mathcal{H}$-DETR~\cite{HDETR} adds an auxiliary one-to-many matching branch to the original one-to-one matching branch.

Our redundancy assignment strategy is tailored for lane topology reasoning, leveraging multiple variants of lanes to enrich valid topology supervision.
It does not require the introduction of additional queries, but instead enhances lane topology performance by providing additional supervision to the existing ones.

%% file: sec/3_method.tex
\section{Method}
\subsection{Overview}
Lane topology reasoning task involves detecting all lanes from multi-view images and traffic elements (\textit{e.g.}, traffic lights and road signs) from the front-view image, and subsequently inferring their topological relationships.
We represent each lane instance as a polyline $\mathbf{L}_p \in \mathbb{R}^{N_P \times 3}$ consisting of $N_P$ uniformly sampled discrete points, where $p = 1, 2, \dots, N_L$, and each traffic element as a bounding box $\mathbf{T}_q \in \mathbb{R}^{4}$, where $q = 1, 2, \dots, N_T$. Here, $N_L$ and $N_T$ denote the total number of lanes and traffic elements, respectively.
Topology reasoning consists of two subtasks, \textit{i.e.}, lane-lane topology and lane-traffic topology prediction. 
Lane-lane topology is represented as a binary matrix $\mathbf{G}^{ll} \in \mathbb{R}^{N_L \times N_L}$, where $\mathbf{G}^{ll}_{pq} = 1$ indicates that lane $\mathbf{L}_p$ is connected to lane $\mathbf{L}_q$, and $0$ otherwise.
Similarly, lane-traffic topology is represented by $\mathbf{G}^{lt} \in \mathbb{R}^{N_L \times N_T}$, where $\mathbf{G}^{lt}_{pq} = 1$ denotes a correspondence between lane $\mathbf{L}_p$ and traffic element $\mathbf{T}_q$.


We present a typical lane topology reasoning pipeline~\cite{TopoNet, TopoMLP, TopoLogic, Topo2D, LaneSegNet, SMERF} in Figure~\ref{fig:pipeline}.
First, multi-view images are processed by an image feature extractor to obtain multi-scale multi-view feature maps using a backbone and neck network (\textit{e.g.}, ResNet-50~\cite{ResNet} and FPN~\cite{FPN}).
Traffic elements are detected by feeding the front-view feature map into a traffic element detector, which consists of a Transformer~\cite{transformer} encoder and a DETR-style~\cite{DETR} object decoder.
For lane detection, following the approach of BEVFormer~\cite{BEVFormer}, we apply a view transformation encoder to the multi-view image features to generate a BEV feature map $\mathbf{F} \in \mathbb{R}^{H \times W \times C}$, which is then processed by a DETR-style~\cite{DETR} lane decoder (detailed in Section~\ref{sec:detr}) and a lane head sequentially.
Here, $H$, $W$ and $C$ denote the height, width and channel number of the BEV feature respectively.
Based on the detection results, MLP-based topology prediction heads (\textit{i.e.}, LT Topo head and LL Topo head) are used to infer both lane-lane and lane-traffic element topological relationships.
Our model-agnostic redundancy assignment strategy is applied to this pipeline by modifying the Transformer~\cite{transformer} layer of lane decoder, which will be elaborated in Section~\ref{sec:ratd}.


\subsection{Standard DETR-Style Lane Decoder}
\label{sec:detr}
As our proposed method builds upon the standard DETR-style~\cite{DETR} lane decoder, we begin by revisiting its basic architecture and supervision scheme to provide a clear background for our method.

\noindent\textbf{Architecture.}
In the standard DETR-style~\cite{DETR} lane decoder, a set of learnable lane queries is defined as input and interacts with the image features to generate refined lane queries through several cascaded Transformer~\cite{transformer} decoder layers. 
Specifically, within each decoder layer, the lane queries are updated as follows:
\begin{gather}
\bar{\mathbf{Q}}^{i-1} = \text{LN}(\text{SelfAttn}^i(\mathbf{Q}^{i-1}) + \mathbf{Q}^{i-1}), \\
\tilde{\mathbf{Q}}^{i-1} = \text{LN}(\text{CrossAttn}^i(\bar{\mathbf{Q}}^{i-1}, \mathbf{F}) + \bar{\mathbf{Q}}^{i-1}), \\
\mathbf{Q}^i = \text{LN}(\text{FFN}^i(\tilde{\mathbf{Q}}^{i-1})) + \tilde{\mathbf{Q}}^{i-1}),
\end{gather}
where $i=1,2,\dots,N$ index the decoder layer. $\mathbf{Q}^{i-1}\in \mathbb{R}^{\bar{N}_L\times C}$ represents the lane queries fed into the $i$-th layer, and $\bar{N}_L$ denotes the number of lane queries.
$\mathbf{F}$ represents the BEV features, LN$(\cdot)$ and FFN$(\cdot)$ denote layer normalization and the feedforward network, respectively.
Two types of attention layers are included in each decoder layer, \textit{i.e.}, self-attention and cross-attention layers.

\noindent\textbf{Detection Supervision.}
The DETR-style~\cite{DETR} lane decoder employs one-to-one assignment in the detection stage, where each ground truth lane is matched to one prediction through bipartite matching:
\begin{gather}
(\bar{y}_{\sigma(1)}, y_1), (\bar{y}_{\sigma(2)}, y_2), ..., (\bar{y}_{\sigma({N_L})}, y_{N_L}),
\end{gather}
where $\sigma$ represents the optimal permutation of $N_L$ indices, $\bar{y}$ and $y$ correspond to the predictions and ground truth lanes.
Detection supervision is then applied by computing the loss between the one-to-one matched predictions and ground-truths.

\noindent\textbf{Topology Supervision.}
After selecting positive predictions using one-to-one assignment during the detection stage, the supervision signal for the predicted topological relationship $\bar{G}^{ll}$ and $\bar{G}^{lt}$ can be constructed based on the ground truth topology $G^{ll}$ and $G^{lt}$ respectively. 
Taking $\bar{G}^{ll}$ as an example, the topological relationships among positive lane samples are aligned with those of the corresponding ground truth, while relationships involving negative samples are assigned a value of 0. The formulation is as follows:
\begin{gather}
Z^{ll}_{rs} = 
\begin{cases}
G^{ll}_{pq}, & \text{if } r=\sigma(p) \text{ and } s=\sigma(q), \\
0, & \text{others,}
\end{cases}
\end{gather}
where $Z^{ll}_{rs}$ is the ground-truth for $\bar{G}^{ll}_{rs}$. 
Topology loss with full supervision in Figure~\ref{fig:teaser}(a) can be then formulated as:
\begin{gather}
\mathcal{L}_{o2o}^{ll}=\sum\limits_{r=1}^{\bar{N}_L}\left(\sum\limits_{s=1}^{\bar{N}_L}\ell_c(\bar{G}^{ll}_{rs}, Z_{rs}^{ll})\right),
\end{gather}
where $\ell_c$ is implemented as a classification loss.
In contrast, topology loss with valid-only supervision in Figure~\ref{fig:teaser}(b) is formulated as:
\begin{gather}
\mathcal{L}_{o2o}^{ll}=\sum\limits_{p=1}^{N_L}\left(\sum\limits_{q=1}^{N_L}\ell_c\left(\bar{G}^{ll}_{\sigma(p)\sigma(q)}, Z_{\sigma(p)\sigma(q)}^{ll}\right)\right).
\label{eq:llloss}
\end{gather}
The supervision for lane-traffic topologies follows the same construction procedure as lane-lane topologies.

\subsection{Lane Decoder with Redundancy Assignment}
\label{sec:ratd}
As shown in Figure~\ref{fig:pipeline}(b), building upon the standard DETR-style~\cite{DETR} lane decoder described above, we introduce three key modifications to each Transformer~\cite{transformer} layer to support redundancy assignment and enhance topology supervision, including: (1) reordered attention layers, (2) auxiliary topology supervision with one-to-many assignment, and (3) multiple parallel cross-attention. 

\noindent\textbf{(1) Reordered attention layers.}
In a standard DETR-style~\cite{DETR} decoder layer, queries first interact with each other through the self-attention (SA) layer, then interact with image features in the cross-attention (CA) layer, and are finally projected by a feedforward network (FFN).
We reorder this sequence to CA $\rightarrow$ SA $\rightarrow$ FFN, allowing each lane query to first independently attend to BEV features before redundancy suppression by the SA layer.
This design mimics the classic RPN + NMS paradigm in traditional object detection~\cite{FasterRCNN}, where multiple candidates are generated before deduplication.
It creates opportunities for one-to-many assignment by preserving redundant yet plausible variants of each ground-truth.

\noindent\textbf{(2) Auxiliary topology supervision with one-to-many assignment.}
By reordering the Transformer~\cite{transformer} decoder, we enable the cross-attention layers to generate redundant yet diverse intermediate lane queries.
As these queries are obtained before self-attention suppression, they serve as plausible geometric variants of the same ground-truth instance, allowing multiple high-quality predictions to be retained per lane.
To exploit this diversity for better topology learning, we introduce auxiliary topology supervision by applying a one-to-many assignment scheme between the intermediate predictions and ground-truth lanes, significantly increasing the quantity and diversity of valid topology supervision signals.
Specifically, within the output of each cross-attention block, we perform one-to-many assignment to identify multiple high-quality predictions for each ground-truth, and the matching results are formulated as:
\begin{gather}
\left(\left\{\bar{y}_{\sigma_k(1)}\right\}_{k=1}^{K}, y_1\right),\dots, \left(\left\{\bar{y}_{\sigma_k(N_L)}\right\}_{k=1}^{K}, y_{N_L}\right),
\end{gather}
where $K$ is the number of matched predictions for each ground-truth lane, the supervisory signal for the corresponding lane-lane topology prediction $\bar{G}^{ll}$ is defined as:
\begin{gather}
Z^{ll}_{rs} = 
\begin{cases}
G^{ll}_{pq}, & \text{if } r\in\left\{\sigma_k(p)\right\}_{k=1}^{K} \text{ and } s\in\left\{\sigma_k(q)\right\}_{k=1}^{K}, \\
0, & \text{others.}
\end{cases}
\end{gather}
Then we apply auxiliary topology loss between the valid part of $\bar{G}^{ll}$ and its corresponding supervision signal $Z^{ll}$ similar to Equation~\ref{eq:llloss}.
In a similar way, we can also apply auxiliary supervision to the topology of lane-traffic elements, yet with one-to-one assignment applied to traffic elements.
It is noted that for the lane queries output by each decoder layer, we still supervise both lane prediction and lane topology prediction using a one-to-one assignment strategy.

\noindent\textbf{(3) Multiple parallel cross-attention.}
Instead of using a single CA block, we further instantiate $M$ parallel CA blocks in each decoder layer, each with independent parameters.
All blocks take the same set of lane queries as input but interact with the BEV feature from different parameter perspectives, thus enabling these queries to generate lane proposals with diverse spatial and semantic patterns.
As illustrated in the Figure~\ref{fig:pipeline}, given lane queries $\mathbf{Q}^{i-1}$ as the input of the $i$-th decoder layer, our parallel CA blocks process them as:
\begin{gather}
\bar{\mathbf{Q}}^{i-1}_m = \text{LN}(\text{CrossAttn}^{i}_m(\mathbf{Q}^{i-1}, \mathbf{F}) + \mathbf{Q}^{i-1}),
\end{gather}
where $m = 1, 2, ..., M$ denotes the index of the CA block. 
Afterwards, we fuse the $M$ sets of intermediate lane queries to form a unified set of queries as follows:
\begin{gather}
\bar{\mathbf{Q}}^{i-1} = \text{Linear}(\text{Concat}(\bar{\mathbf{Q}}^{i-1}_1, \bar{\mathbf{Q}}^{i-1}_2, ..., \bar{\mathbf{Q}}^{i-1}_M)),
\end{gather}
where Concat$(\cdot)$ denotes concatenating queries at the same index along channel dimension and Linear$(\cdot)$ represents a linear layer to project the concatenated features back to the original dimensions.
Auxiliary topology losses calculated from different parallel cross-attention blocks are accumulated to obtain the final auxiliary topology supervision $\mathcal{L}^{topo}_{o2m}$.

\begin{table*}[t]
  \caption{Comparison results of lane centerline perception and reasoning on OpenLane-V2~\cite{OpenLaneV2}.
  All models are trained for 24 epochs with a ResNet-50~\cite{ResNet} backbone.
  $*$ indicates reimplemented results based on the released codebase of TopoMLP~\cite{TopoMLP}.
  }
  \label{tab:sota}
  \centering
  \begin{tabular}{c|c|c|c|ccccc}
    \thickhline
    Data & Method & Venues & w/ RATopo & $\text{TOP}_{ll}\uparrow$ & $\text{TOP}_{lt}\uparrow$ & $\text{DET}_l\uparrow$ & $\text{DET}_t\uparrow$ & $\text{OLS}\uparrow$ \\
    \hline \hline
    \multirow{9}{*}{\textit{subset\_A}} & STSU~\cite{STSU} & ICCV 2021 &  & 2.9 & 19.8 & 12.7 & 43.0 & 29.3 \\
    & VectorMapNet~\cite{VectorMapNet} & ICML 2023 & &  2.7 & 9.2 & 11.1 & 41.7 & 24.9 \\
    & MapTR~\cite{MapTR} & ICLR 2023 & & 5.9 & 15.1 & 17.7 & 43.5 & 31.0 \\
    \cline{2-9}
    & TopoNet~\cite{TopoNet} & \multirow{2}{*}{Arxiv 2023} & & 10.9 & 23.8 & 28.6 & 48.6 & 39.8 \\
    & TopoNet~\cite{TopoNet} & & \checkmark & 29.5 \textcolor{myblue}{(+18.6)} & 32.6 \textcolor{myblue}{(+8.8)} & 29.6 \textcolor{myblue}{(+1.0)} & 49.0 \textcolor{myblue}{(+0.4)} & 47.5 \textcolor{myblue}{(+7.7)} \\
    \cline{2-9}
    & TopoMLP~\cite{TopoMLP} & \multirow{2}{*}{ICLR 2024} & & 21.7 & 26.9 & 28.5 & 49.5 & 44.1 \\
    & TopoMLP~\cite{TopoMLP} & & \checkmark & 24.0 \textcolor{myblue}{(+2.3)} & 30.0 \textcolor{myblue}{(+3.1)} & 28.6 \textcolor{myblue}{(+0.1)} & \textbf{51.6} \textcolor{myblue}{(+2.1)} & 46.0 \textcolor{myblue}{(+1.9)} \\
    \cline{2-9}
    & TopoLogic~\cite{TopoLogic} & \multirow{2}{*}{NeurIPS 2024} & & 23.9 & 25.4 & 29.9 & 47.2 & 44.1 \\
    & TopoLogic~\cite{TopoLogic} & & \checkmark &  \textbf{32.2} \textcolor{myblue}{(+8.3)} & \textbf{33.9} \textcolor{myblue}{(+ 8.5)} & \textbf{31.8} \textcolor{myblue}{(+1.9)} & 49.4 \textcolor{myblue}{(+2.2)} & \textbf{49.0} \textcolor{myblue}{(+4.9)} \\
    \hline \hline
    \multirow{6}{*}{\textit{subset\_B}} & TopoNet~\cite{TopoNet} & \multirow{2}{*}{Arxiv 2023} & & 6.7 & 16.7 & 24.4 & 52.6 & 36.0 \\
    & TopoNet~\cite{TopoNet} & & \checkmark & 33.3 \textcolor{myblue}{(+26.6)} & 25.8 \textcolor{myblue}{(+9.1)} & 29.7 \textcolor{myblue}{(+5.3)} & 55.1 \textcolor{myblue}{(+2.5)} & 48.3 \textcolor{myblue}{(+12.3)} \\
    \cline{2-9}
    & TopoMLP$*$~\cite{TopoMLP} & \multirow{2}{*}{ICLR 2024} & & 20.8 & 20.3 & 21.6 & 59.1 & 42.9 \\
    & TopoMLP~\cite{TopoMLP} & & \checkmark & 27.5 \textcolor{myblue}{(+6.7)} & 25.2 \textcolor{myblue}{(+4.9)} & 25.0 \textcolor{myblue}{(+3.4)} & \textbf{60.5} \textcolor{myblue}{(+1.4)} & 47.0 \textcolor{myblue}{(+4.1)} \\
    \cline{2-9}
    & TopoLogic~\cite{TopoLogic} & \multirow{2}{*}{NeurIPS 2024} & & 21.6 & 17.9 & 25.9 & 54.7 & 42.3 \\
    & TopoLogic~\cite{TopoLogic} & & \checkmark & \textbf{37.3} \textcolor{myblue}{(+15.7)} & \textbf{27.4} \textcolor{myblue}{(+9.5)} & \textbf{33.1} \textcolor{myblue}{(+7.2)} & 54.7 \textcolor{myblue}{(+0.0)} & \textbf{50.3} \textcolor{myblue}{(+8.0)} \\
    \hline
  \end{tabular}
\end{table*}

\begin{table}[t]
  \caption{
  Comparison results of lane segment perception and reasoning on the OpenLane-V2~\cite{OpenLaneV2} \textit{subset\_A}. 
  All experiments are conducted using ResNet-50~\cite{ResNet} as the backbone, and are also trained for 24 epochs.
  }
  \label{tab:laneseg}
  \tabcolsep=0.06cm
  \centering
  \begin{tabular}{c|c|cccc}
    \thickhline
    Method & Venues & $\text{TOP}_{lsls}\uparrow$ & $\text{mAP}\uparrow$ & $\text{AP}_{ls}\uparrow$ & $\text{AP}_{ped}\uparrow$ \\
    \hline \hline
    TopoNet~\cite{TopoNet} & Arxiv 2023 & - & 23.0 & 23.9 & 22.0 \\
    MapTR~\cite{MapTR} & ICLR 2023 & - & 27.0 & 25.9 & 28.1 \\
    MapTRv2~\cite{MapTRv2} & IJCV 2024 & - & 28.5 & 26.6 & 30.4 \\
    LaneSegNet~\cite{LaneSegNet} & ICLR 2024 & 25.4 & 32.6 & 32.3 & 32.9 \\
    \hline
    LaneSegNet~\cite{LaneSegNet} & \multirow{2}{*}{-} & \textbf{31.3} & \textbf{33.3} & \textbf{33.1} & \textbf{33.4} \\
    \textcolor{myblue}{+ RATopo} & & \textcolor{myblue} {(+5.9)} & \textcolor{myblue} {(+0.7)} & \textcolor{myblue} {(+0.8)} & \textcolor{myblue} {(+0.5)} \\
    \hline
  \end{tabular}
\end{table}

\begin{table}[t]
  \caption{Comparison between different one-to-many assignment methods. TopoLogic~\cite{TopoLogic} with valid-only supervision serves as the baseline method.
  \textit{Naive O2M} denotes naive one-to-many assignment solution.
  \textit{Group O2M} denotes the one-to-many assignment strategy proposed by Group DETR~\cite{GroupDETR}.
  }
  \label{tab:one2many}
  \tabcolsep=0.06cm
  \centering
  \begin{tabular}{c|ccccc}
    \thickhline
    O2M method & $\text{TOP}_{ll}\uparrow$ & $\text{TOP}_{lt}\uparrow$ & $\text{DET}_l\uparrow$ & $\text{DET}_t\uparrow$ & $\text{OLS}\uparrow$ \\
    \hline \hline
    w/o O2M & 27.2 & 28.2 & 30.1 & 50.0 & 46.4 \\
    Naive O2M & 27.0 & 28.6 & 29.9 & \textbf{50.2} & 46.4 \\
    Group O2M~\cite{GroupDETR} & 29.8 & 30.3 & 31.3 & 48.6 & 47.4 \\
    Our RATopo & \textbf{32.2} & \textbf{33.9} & \textbf{31.8} & 49.4 & \textbf{49.0} \\
    \hline
  \end{tabular}
\end{table}

\subsection{Loss Function}
We define the overall loss function of our method as follows:
\begin{gather}
\mathcal{L} = \mathcal{L}^{det}_{o2o} + \mathcal{L}^{topo}_{o2o} + \lambda_{o2m}\mathcal{L}^{topo}_{o2m},  
\end{gather}
where $\mathcal{L}^{det}_{o2o}$, $\mathcal{L}^{topo}_{o2o}$, and $\mathcal{L}^{topo}_{o2m}$ represent the detection loss with one-to-one assignment, the topology loss with one-to-one assignment, and the topology loss with one-to-many assignment, respectively.
Specifically, the detection loss $\mathcal{L}^{det}_{o2o}$ consists of two parts:
\begin{gather}
\mathcal{L}^{det}_{o2o} = \lambda^l \mathcal{L}^{l}_{o2o} + \lambda^t \mathcal{L}^{t}_{o2o}.
\end{gather}
The lane detection loss $\mathcal{L}^{l}_{o2o}$ includes a focal loss~\cite{FocalLoss} for classification, and an L1 loss for regression. In addition to the classification and regression losses, the traffic element loss $\mathcal{L}^{t}_{o2o}$ also incorporates the GIoU loss~\cite{GIoU} to enhance performance.
For topology reasoning, both $\mathcal{L}^{topo}_{o2o}$ and $\mathcal{L}^{topo}_{o2m}$ incorporate supervision for lane-lane topology prediction and lane-traffic element prediction:
\begin{gather}
\mathcal{L}^{topo}_{o2o} = \lambda^{ll}\mathcal{L}^{ll}_{o2o} + \lambda^{lt}\mathcal{L}^{lt}_{o2o}, \\
\mathcal{L}^{topo}_{o2m} = \lambda^{ll}\mathcal{L}^{ll}_{o2m} + \lambda^{lt}\mathcal{L}^{lt}_{o2m}.
\end{gather}
We adopt the focal loss~\cite{FocalLoss} for all topology losses.

%% file: sec/4_experiments.tex
\section{Experiments}

\begin{table*}[t]
  \caption{Effectiveness of different modules.
  We use TopoLogic~\cite{TopoLogic} as the baseline. 
  \textit{Traffic O2M} denotes the incorporation of redundancy assignment strategy for the traffic element detector.
  }
  \label{tab:module}
  \centering
  \begin{tabular}{ccccc|ccccc}
    \thickhline
    Valid-only Sup. & Reorder & Lane O2M & Multi-CA & Traffic O2M & $\text{TOP}_{ll}\uparrow$ & $\text{TOP}_{lt}\uparrow$ & $\text{DET}_l\uparrow$ & $\text{DET}_t\uparrow$ & $\text{OLS}\uparrow$ \\
    \hline \hline
    & & & & & 23.9 & 25.4 & 29.9 & 47.2 & 44.1 \\
    \checkmark & & & & & 27.2 & 28.2 & 30.1 & 50.0 & 46.4 \\
    \checkmark & \checkmark & & & & 28.2 & 27.9 & 30.8 & 50.0 & 46.7 \\
    \checkmark & \checkmark & \checkmark & & & 30.5 & 31.5 & 30.3 & 50.3 & 48.0 \\
    \checkmark & \checkmark & & \checkmark & & 29.8 & 29.1 & 29.7 & \textbf{50.5} & 47.2 \\
    \checkmark & \checkmark & \checkmark & \checkmark & & \textbf{32.2} & \textbf{33.9} & \textbf{31.8} & 49.4 & \textbf{49.0} \\
    \checkmark & \checkmark & \checkmark & \checkmark & \checkmark & 31.9 & 32.7 & 31.0 & 49.5 & 48.5 \\
    \hline
  \end{tabular}
\end{table*}

\begin{table}[t]
  \caption{Number of positive samples assigned per lane.}
  \label{tab:posnum}
  \tabcolsep=0.08cm
  \centering
  \begin{tabular}{c|ccccc}
    \thickhline
    Number & $\text{TOP}_{ll}\uparrow$ & $\text{TOP}_{lt}\uparrow$ & $\text{DET}_l\uparrow$ & $\text{DET}_t\uparrow$ & $\text{OLS}\uparrow$ \\
    \hline \hline
    1 & 30.4 & 31.0 & 29.5 & 50.1 & 47.6 \\
    2 & \textbf{32.2} & 32.2 & 29.9 & 50.2 & 48.4 \\
    3 & \textbf{32.2} & \textbf{33.9} & \textbf{31.8} & 49.4 & \textbf{49.0} \\
    4 & 31.7 & 31.9 & 31.2 & 49.5 & 48.4 \\
    5 & 30.8 & 32.6 & 29.8 & \textbf{50.6} & 48.3 \\
    \hline
  \end{tabular}
\end{table}

\begin{table}[t]
  \caption{Number of parallel cross-attention blocks.}
  \label{tab:canum}
  \tabcolsep=0.08cm
  \centering
  \begin{tabular}{c|ccccc}
    \thickhline
    Number & $\text{TOP}_{ll}\uparrow$ & $\text{TOP}_{lt}\uparrow$ & $\text{DET}_l\uparrow$ & $\text{DET}_t\uparrow$ & $\text{OLS}\uparrow$ \\
    \hline \hline
    1 & 30.5 & 31.5 & 30.3 & 50.3 & 48.0 \\
    2 & 30.8 & 31.2 & 31.1 & \textbf{50.9} & 48.3 \\
    4 & 32.2 & \textbf{33.9} & \textbf{31.8} & 49.4 & \textbf{49.0} \\
    6 & \textbf{32.3} & 33.7 & 31.4 & 48.2 & 48.6 \\
    \hline
  \end{tabular}
  \vspace{-4mm}
\end{table}

\subsection{Datasets and Metrics}
We evaluate our RATopo on the OpenLane-V2~\cite{OpenLaneV2} benchmark, which is developed based on the Argoverse2~\cite{Argoverse2} and nuScenes~\cite{nuScenes} datasets. 
The OpenLane-V2~\cite{OpenLaneV2} benchmark consists of two subsets: \textit{subset\_A} and \textit{subset\_B}, each containing 1,000 scenes with multi-view image sequences sampled at 2Hz.
The annotation comprises four components: lanes, traffic elements, lane-lane topology relationship, and lane-traffic element topology relationship. 

In the task of lane centerline perception and reasoning, a set of evaluation metrics is defined, including $\text{DET}_l$, $\text{DET}_t$, $\text{TOP}_{ll}$, and $\text{TOP}_{lt}$. 
The former two evaluate the detection performance of lanes and traffic elements respectively.
Meanwhile, the metrics $\text{TOP}_{ll}$ and $\text{TOP}_{lt}$ are employed to assess the topology reasoning performance regarding lane-lane and lane-traffic element relationships, respectively.
The overall OpenLane-V2 Score (OLS) is calculated as follows:
\begin{gather}
\text{OLS} = \frac{1}{4} \left( \text{DET}_l+\text{DET}_t+\sqrt{\text{TOP}_{ll}}+\sqrt{\text{TOP}_{lt}} \right).
\end{gather}

For lane segment perception and reasoning, the evaluation metrics include average precision metrics $\text{AP}_{ls}$ and $\text{AP}_{ped}$, along with the mAP computed as the average of $\text{AP}_{ls}$ and $\text{AP}_{ped}$. 
Additionally, it incorporates the topology metric $\text{TOP}_{lsls}$.

\subsection{Implementation Details}
Following previous methods, we adopt a ResNet-50~\cite{ResNet} as the backbone and use an FPN~\cite{FPN} as the neck to extract image features at downsampling scales of $1/4$, $1/8$, $1/16$, and $1/32$.
We utilize a view transformation encoder with 3 Transformer~\cite{transformer} layers, as proposed by BEVFormer~\cite{BEVFormer}. 
The size of the BEV features is set to $200\times 100$. 
The query numbers for lane centerlines and traffic elements are set to 300 and 100, respectively. 
The values for the loss weights $\lambda^{l}$, $\lambda^{t}$, $\lambda^{ll}$, and $\lambda^{lt}$ are set to 1.0, 1.0, 5.0, and 5.0, respectively. 
The weight $\lambda_{o2m}$ for auxiliary one-to-many topology supervision is set to 2.0.
All images are resized to match the dimensions of $1024\times 775$ pixels. 
The training is performed using the AdamW~\cite{adamw} optimizer with an initial learning rate of $2\times 10^{-4}$. 
The batch size is set to 8. 
All experiments are conducted on 8 Tesla A100 GPUs.

\subsection{Comparison on OpenLane-V2 Dataset}
Table~\ref{tab:sota} provides a comparison with state-of-the-art methods on OpenLane-V2 dataset.
Owing to its model-agnostic nature, our RATopo strategy integrates seamlessly with state-of-the-art open-source frameworks like TopoNet, TopoMLP, and TopoLogic and consistently delivers significant gains on the core metrics for topology reasoning, \textit{i.e.}, $\text{TOP}_{ll}$ and $\text{TOP}_{lt}$.
For example, On \textit{subset\_A}, our RATopo
boosts $\text{TOP}_{ll}$ by 18.6\% and 8.3\% for TopoNet and TopoLogic, while also improving their $\text{TOP}_{lt}$ metric by 8.8\% and 8.5\%.
Notably, TopoLogic with our RATopo achieved 32.2\% on $\text{TOP}_{ll}$, and 33.9\% on $\text{TOP}_{lt}$, setting a new state-of-the-art performance. 
On \textit{subset\_B}, we observe consistent improvements across two topology-related metrics by integrating our RATopo strategy. 
These experimental results thoroughly validate the effectiveness and generalization capability of our method.

We also implement our RATopo strategy on LaneSegNet. 
As shown in Table~\ref{tab:laneseg}, our method brings a 5.9\% improvement to the $\text{TOP}_{lsls}$ metric, proving the generalization capability of our method across different topology reasoning tasks.

\begin{figure*}[t]
  \centering
  \begin{subfigure}{0.45\linewidth}
    \centering
    \includegraphics[width=\linewidth]{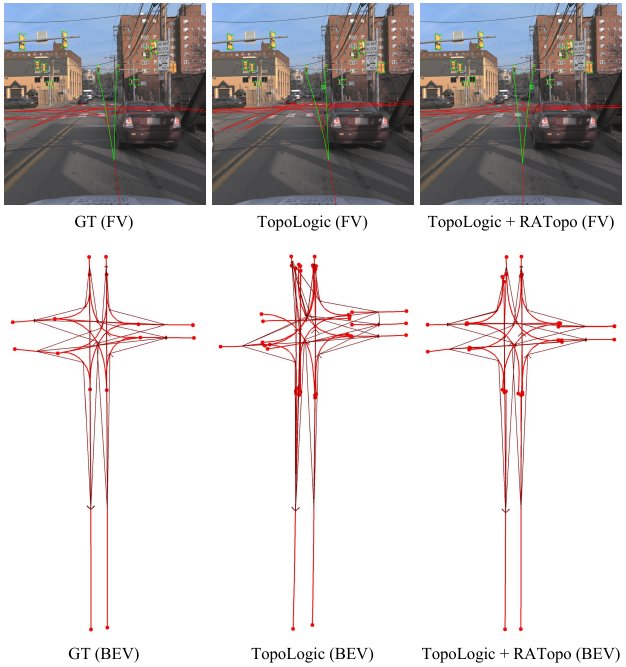}
    \caption{}
    \label{fig:subset_a_1}
  \end{subfigure}
  \begin{subfigure}{0.45\linewidth}
    \centering
    \includegraphics[width=\linewidth]{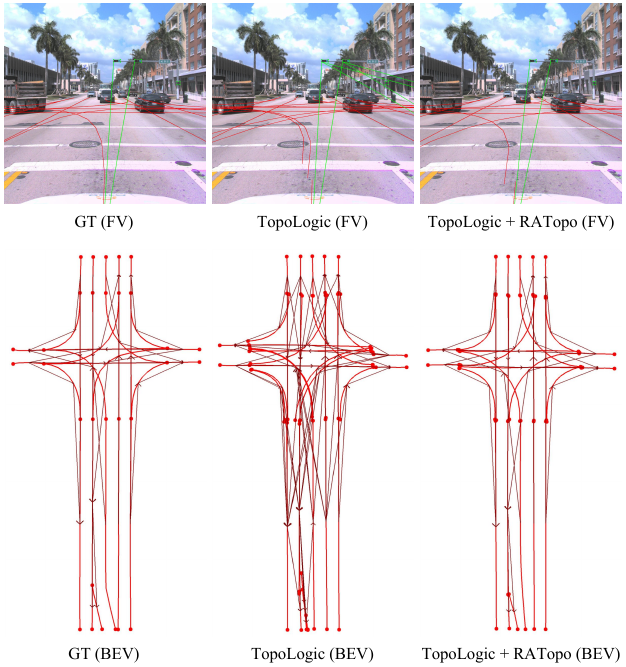}
    \caption{}
    \label{fig:subset_a_2}
  \end{subfigure}
  \begin{subfigure}{0.45\linewidth}
    \centering
    \includegraphics[width=\linewidth]{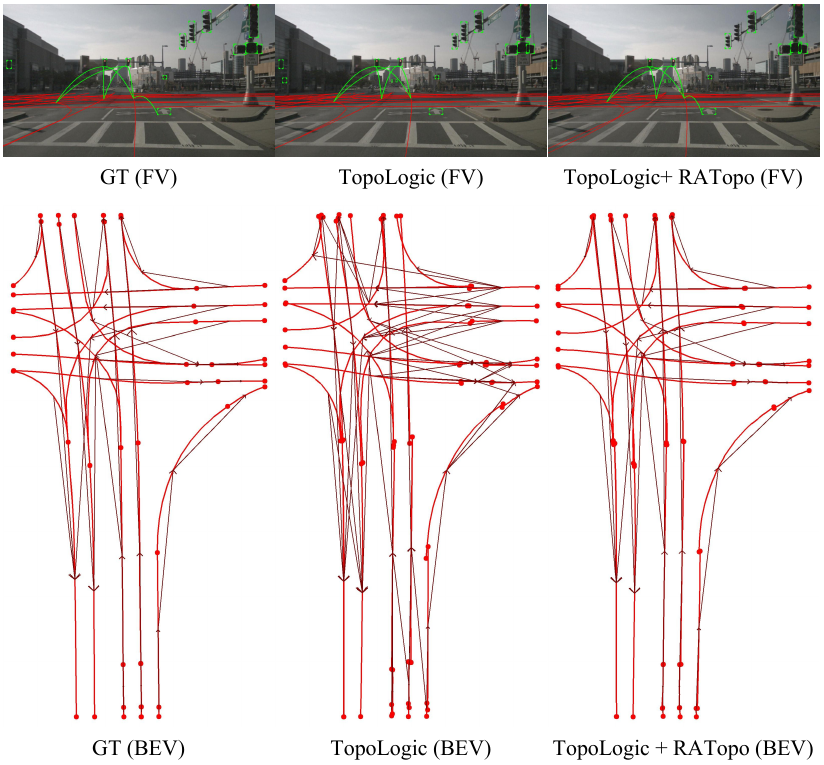}
    \caption{}
    \label{fig:subset_b_1}
  \end{subfigure}
  \begin{subfigure}{0.45\linewidth}
    \centering
    \includegraphics[width=\linewidth]{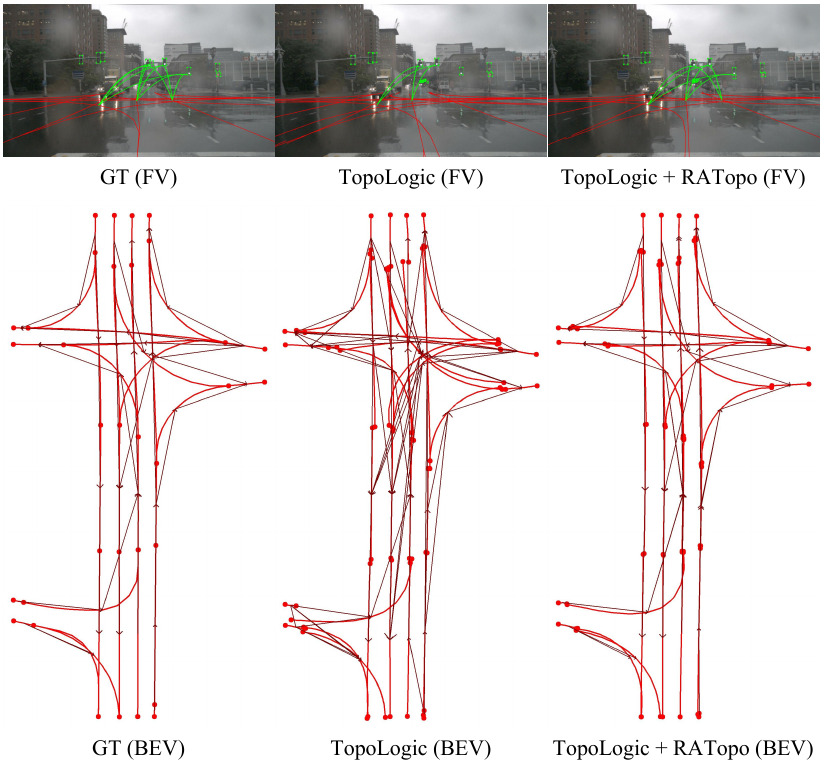}
    \caption{}
    \label{fig:subset_b_2}
  \end{subfigure}
  \caption{Qualitative comparison between TopoLogic w/o and w/ our RATopo strategy on OpenLane-V2~\cite{OpenLaneV2} dataset. (a)(b) Results on \textit{subset\_A}. (c)(d) Results on \textit{subset\_B}.}
  \Description{}
  \label{fig:vis_openlanev2}
\end{figure*}

\subsection{Ablation Study}
All ablation experiments are conducted OpenLane-V2 \textit{subset\_A} with TopoLogic as baseline.
All the models are trained for 24 epochs with ResNet-50 as the backbone.


\noindent \textbf{Comparison with other one-to-many assignment methods.}
We compare our redundancy assignment strategy with other one-to-many assignment methods in Table~\ref {tab:one2many}.
As shown in the 2nd row, naively applying one-to-many assignment without reordering the self-attention and cross-attention layers brings no improvement over the baseline, due to the inherent redundancy suppression in the standard Transformer-based lane decoder that limits the diversity of predicted lanes.
We also implement the one-to-many assignment strategy proposed in Group DETR~\cite{GroupDETR}, where additional multiple groups of lane queries are defined to increase the quantity of valid topology supervision.
As shown in the third row, this strategy yields moderate improvements over the baseline, which also supports our claim that \textit{incorporating more diverse lane geometries can facilitate better topology learning}.
However, since the auxiliary supervision is applied exclusively to the newly added query groups, the original set of queries remains sparsely supervised.
In contrast, our RATopo strategy directly enhances the diversity of original queries, enabling a greater proportion of them to participate in topology learning, which leads to stronger performance gains.

\noindent \textbf{Component analysis.}
We include the ablation results of different components of our RATopo strategy in Table~\ref{tab:module}.
The original TopoLogic result (1st row) reported in its paper under full supervision, which suffers from a severe dominance of invalid topologies and yields suboptimal performance. 
Therefore, we re-implement this method using valid-only supervision (2nd row) to provide a more solid baseline for our ablation study.
While reordering the self-attention and cross-attention layers alone yields minimal performance change (3rd row), introducing auxiliary topology supervision via one-to-many assignment on the intermediate lane queries leads to notable improvements (\textit{i.e.}, 2.3\% for $\text{TOP}_{ll}$ and 3.6\% for $\text{TOP}_{lt}$). 
This demonstrates the effectiveness of enriching valid topology supervision through diverse one-to-many associations, without adding parameters or inference-time overhead.
Introducing multiple parallel cross-attention blocks alone also improves topology reasoning performance (4th row), highlighting their effectiveness in generating lane queries with diverse spatial and semantic patterns.
Combining these two components to form our full RATopo strategy results in further improvements, as demonstrated in the second-to-last row of Table~\ref{tab:module}.
We also implement the one-to-many assignment scheme for traffic element detection, but observe a performance drop in the last row. This further supports our claim that the effectiveness of the RATopo strategy lies in capturing diverse lane geometries to facilitate the learning of topological relationships.


\noindent \textbf{Number of positive samples.}
In Table~\ref{tab:posnum}, we conduct ablation experiments to determine the optimal number of positive samples assigned to each ground-truth lane under the one-to-many assignment scheme.
Assigning too few positives may result in insufficient valid topology supervision, while assigning too many can introduce numerous low-quality lane predictions, ultimately degrading topology prediction performance.
We find that assigning 3 positive samples per ground-truth lane yields the best overall performance.

\noindent \textbf{Number of parallel cross-attention blocks.}
As shown in Table~\ref{tab:canum}, we evaluate the impact of the number of cross-attention blocks on performance. Our experiments demonstrate that gradually increasing the number consistently improves the topology-related metrics, while also resulting in an increase in model parameters and computational cost. Therefore, we set the number of parallel cross-attention blocks to 4 to balance performance and efficiency.

\subsection{Qualitative Results}


We present qualitative comparison results between TopoLogic w/o and w/ our RATopo strategy in Figure~\ref{fig:vis_openlanev2}.
As shown in Figure~\ref{fig:vis_openlanev2}(a), with our RATopo, Topologic detects more complete and accurate lanes while correctly inferring the topological relationships between them. 
In Figure~\ref{fig:vis_openlanev2}(b), although both methods are able to identify full lane structures, w/ our RATopo performs better in terms of topology reasoning between lanes.
Furthermore, Figure~\ref{fig:vis_openlanev2}(c) and (d) illustrate that RATopo significantly reduces false lane-traffic element topologies.

%% file: sec/5_conclusion.tex
\section{Conclusion}
Lane topology reasoning plays a critical role in modern advanced driver assistance systems.
Motivated by the sparsity of valid topology supervision in existing methods, where supervision signals are derived from one-to-one assignments in the detection stage, we propose RATopo, a model-agnostic redundancy assignment strategy for lane topology reasoning. 
By restructuring the Transformer decoder to enable effective one-to-many assignment and introducing parallel cross-attention blocks, RATopo enables quantity-rich and geometry-diverse topology supervision.
Extensive experiments on the OpenLane-V2 benchmark demonstrate that RATopo consistently improves the performance of various topology reasoning frameworks, validating its effectiveness and generality.
We hope our work can provide new insights into lane topology reasoning.